\documentclass{article}

\usepackage{microtype}
\usepackage{graphicx}
\usepackage{subcaption}
\usepackage{booktabs} % for professional tables

\usepackage{hyperref}

\usepackage[preprint]{icml2026}

\usepackage{amsmath}
\usepackage{amssymb}
\usepackage{mathtools}
\usepackage{amsthm}

\usepackage[capitalize,noabbrev]{cleveref}

\definecolor{r}{RGB}{255,0,0}
\definecolor{g}{RGB}{0,255,0}
\definecolor{b}{RGB}{0,0,255}

\usepackage[normalem]{ulem}  % 提供 \sout 等删除线功能

\theoremstyle{plain}

\theoremstyle{definition}

\theoremstyle{remark}

\usepackage[textsize=tiny]{todonotes}

\icmltitlerunning{Submission and Formatting Instructions for ICML 2026}

\begin{document}

\twocolumn[
  \icmltitle{MoEMambaMIL: Structure-Aware Selective State Space Modeling  \\
    for Whole-Slide Image Analysis}



  \icmlsetsymbol{equal}{*}

  \begin{icmlauthorlist}
    \icmlauthor{Dongqing Xie}{equal,tongji}
    \icmlauthor{Yonghuang Wu}{equal,fudan}
    % \icmlauthor{Firstname3 Lastname3}{comp}
  \end{icmlauthorlist}

  \icmlaffiliation{tongji}{Tongji University}
  \icmlaffiliation{fudan}{Fudan University}
%   \icmlaffiliation{sch}{School of ZZZ, Institute of WWW, Location, Country}

  \icmlcorrespondingauthor{Yonghuang Wu}{yonghuangwu21@m.fudan.edu.cn}
%   \icmlcorrespondingauthor{Firstname2 Lastname2}{first2.last2@www.uk}

  \icmlkeywords{Machine Learning, ICML}

  \vskip 0.3in
]



\printAffiliationsAndNotice{}  % no special notice (required even if empty)

\begin{abstract}

Whole-slide image (WSI) analysis is challenging due to the gigapixel scale of slides and their inherent hierarchical multi-resolution structure. Existing multiple instance learning (MIL) approaches often model WSIs as unordered collections of patches, which limits their ability to capture structured dependencies between global tissue organization and local cellular patterns. Although recent State Space Models (SSMs) enable efficient modeling of long sequences, how to structure WSI tokens to fully exploit their spatial hierarchy remains an open problem.

We propose MoEMambaMIL, a structure-aware SSM framework for WSI analysis that integrates region-nested selective scanning with mixture-of-experts (MoE) modeling. Leveraging multi-resolution preprocessing, MoEMambaMIL organizes patch tokens into region-aware sequences that preserve spatial containment across resolutions. On top of this structured sequence, we decouple resolution-aware encoding and region-adaptive contextual modeling via a combination of static, resolution-specific experts and dynamic sparse experts with learned routing. This design enables efficient long-sequence modeling while promoting expert specialization across heterogeneous diagnostic patterns. Experiments demonstrate that MoEMambaMIL achieves the best performance across 9 downstream tasks. 

\end{abstract}

\section{Introduction}

\begin{figure}
    \centering
    \includegraphics[width=0.9\linewidth]{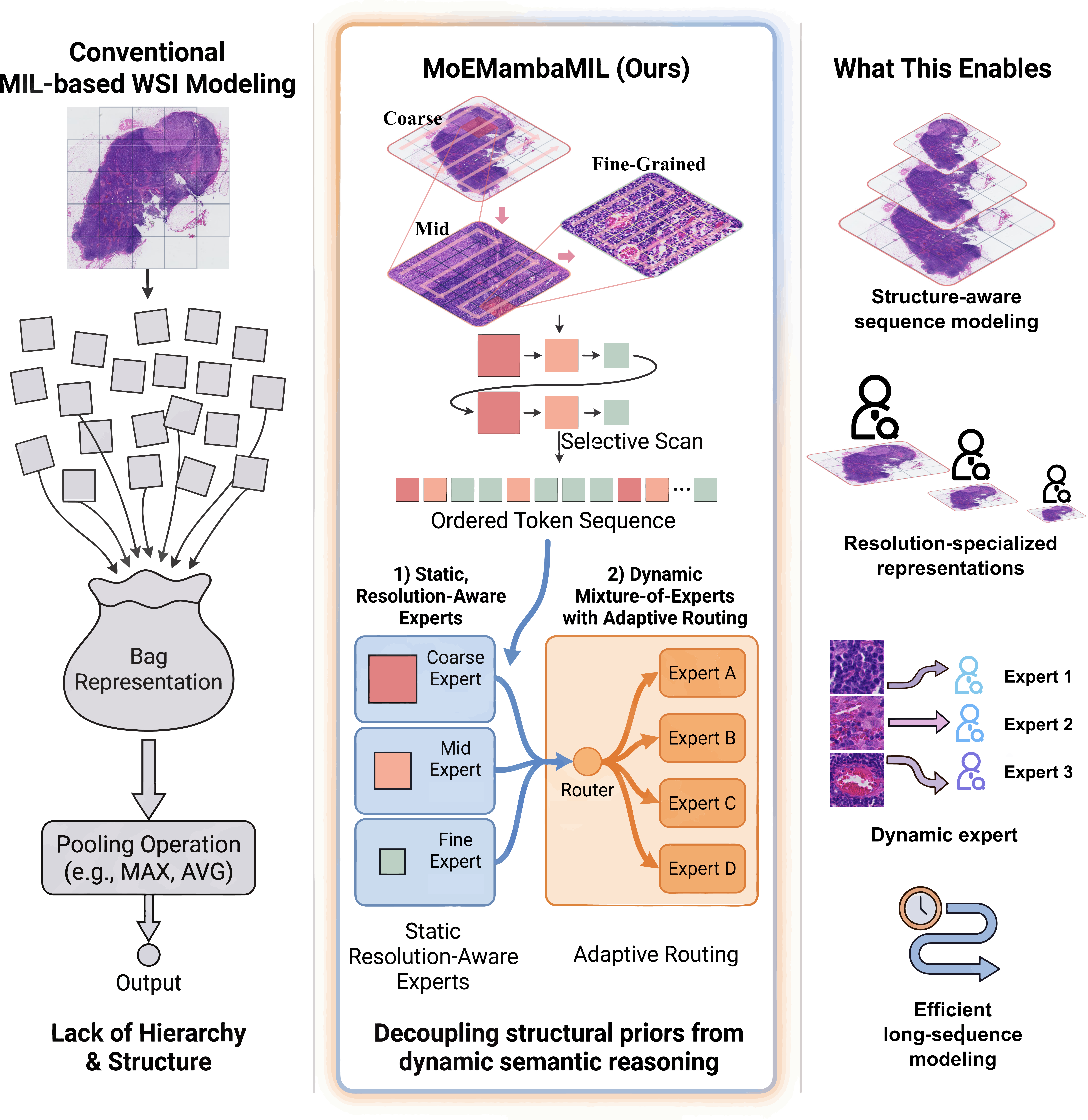}
    \caption{Conceptual comparison between conventional MIL-based WSI modeling and the proposed MoEMambaMIL framework.}
    \label{fig:inno}
\end{figure}

\begin{figure*}[ht]
    \centering
    \includegraphics[width=0.8\linewidth]{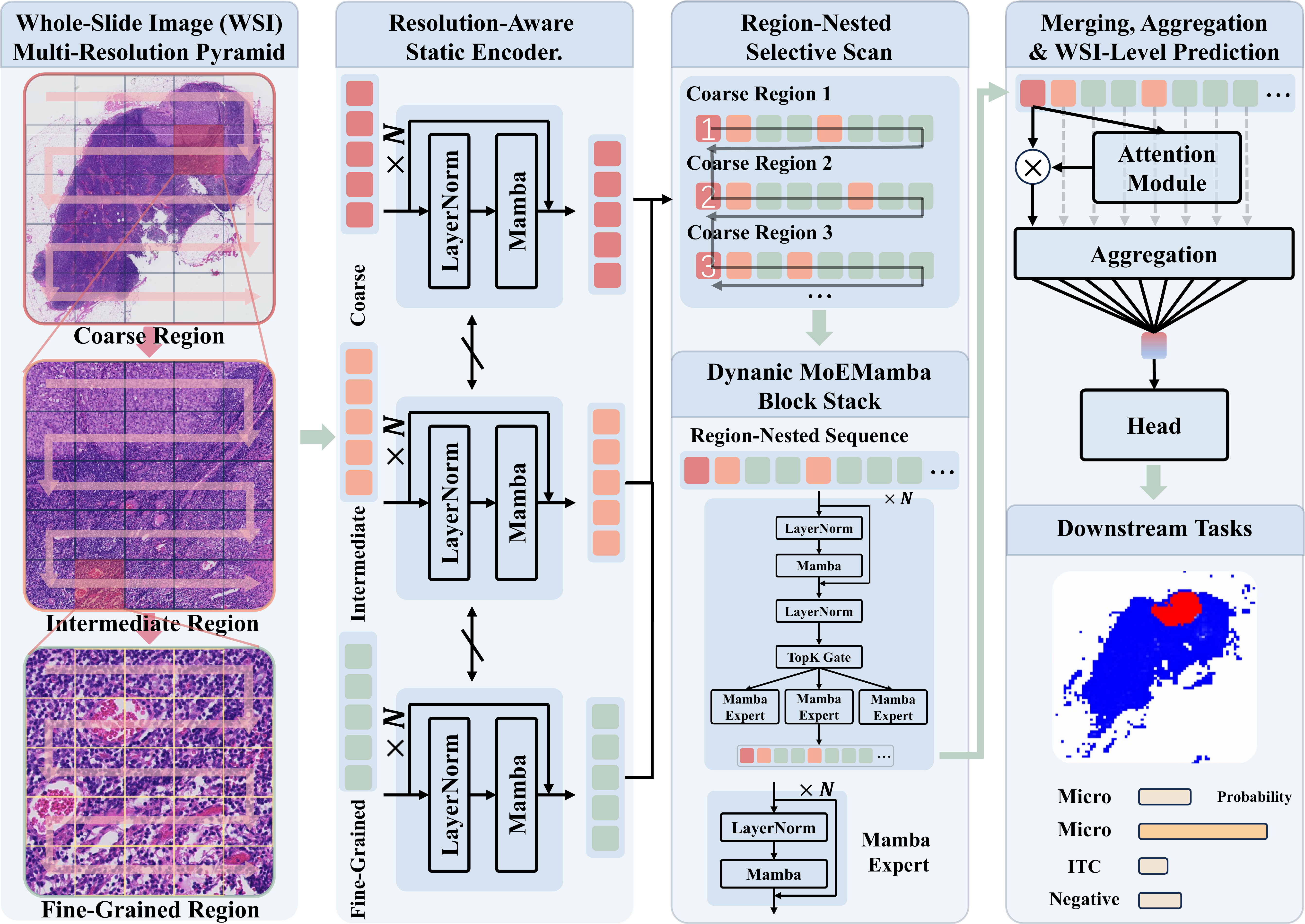}
    \caption{
    Overview of the proposed framework. Multi-resolution WSI patches are first organized into resolution-aware sequences and modeled by static experts to capture structural representations. A selective scan strategy is then applied to construct region-nested scans, yielding region-nested token sequences. These sequences are processed by a MoEMamba backbone with gating and routing mechanisms that dynamically dispatch tokens to Mamba experts. Finally, an attention-based MIL head aggregates token features for WSI-level prediction.
}
    \label{fig:method}
\end{figure*}

Whole-slide images (WSIs) capture diagnostic patterns across multiple spatial scales, but their gigapixel resolution necessitates Multiple Instance Learning (MIL), where thousands of patches are aggregated into a slide-level prediction. While attention-based MIL and Vision Transformers model inter-patch dependencies, their quadratic complexity limits scalability, and many methods still treat patches as an unordered set, ignoring the hierarchical and spatial organization of WSIs.

Structured State Space Models (SSMs), such as Mamba, enable linear-time long-sequence modeling, offering a promising alternative to attention. However, WSIs are inherently two-dimensional and multi-resolution, whereas SSMs operate on one-dimensional sequences. Naïvely flattening patches destroys spatial locality and obscures the hierarchical containment relationships essential for pathological interpretation.

We propose MoEMambaMIL, a structure-aware SSM framework for WSI analysis that explicitly encodes hierarchical and spatial priors. Using standard multi-resolution preprocessing, we introduce a region-nested selective scan that linearizes WSIs by recursively expanding each coarse region together with its higher-resolution descendants, producing a 1D sequence in which patches from the same anatomical region form contiguous subsequences.

On top of this scanned sequence, MoEMambaMIL decouples resolution-specific encoding from region-adaptive contextual modeling via two complementary forms of expert specialization. Static resolution experts are deterministically assigned based on patch resolution to capture scale-specific features, while dynamic sparse experts use learned token-level routing to adaptively model heterogeneous tissue semantics across regions.

By combining region-nested selective scanning with static and dynamic expert specialization, MoEMambaMIL enables efficient long-sequence modeling with linear complexity while respecting the structural organization of WSIs.

Our main contributions are:
\begin{itemize}
\item Region-nested selective scanning, a structure-aware serialization of multi-resolution WSIs for state-space modeling.
\item MoEMambaMIL, a novel MIL framework that decouples resolution-aware encoding from region-adaptive contextual modeling via static and dynamic experts.
\item State-of-the-art performance on multiple WSI benchmarks with linear computational complexity.
\end{itemize}

\section{Related Work}

\subsection{Multiple Instance Learning for Whole-Slide Images}

Whole-slide image (WSI) analysis is commonly formulated as a Multiple Instance Learning (MIL) problem, where each slide is represented as a bag of patch-level instances. Attention-based methods such as AB-MIL~\cite{ilse2018attention} and CLAM~\cite{lu2021clam} enabled weakly supervised localization via learnable pooling.

Subsequent work introduced contextual and multi-scale modeling. TransMIL~\cite{shao2021transmil} applies self-attention to capture global dependencies but suffers from quadratic complexity, while HIPT~\cite{chen2022scaling} adopts a hierarchical design across resolutions. Other methods, including LESS~\cite{zhao2024less}, DGR-MIL~\cite{zhu2024dgrmil}, and PSA-MIL~\cite{castro2026probabilistic}, further enhance scale diversity or spatial reasoning.

However, most approaches treat patches as unordered sets or rely on weak positional cues, failing to explicitly model the hierarchical organization of tissue. In contrast, our method reorganizes multi-resolution patches into a \emph{region-nested sequence}, explicitly preserving spatial containment and biological hierarchy for more semantically meaningful sequence modeling.

\subsection{State Space Models in Vision}

State Space Models (SSMs) have emerged as efficient alternatives to Transformers for long-sequence modeling. S4~\cite{gu2024mamba} introduced structured state transitions with favorable complexity, while Mamba~\cite{gu2024mamba} further enables input-conditioned selective updates, achieving linear-time modeling of long-range dependencies.  
  
Vision adaptations such as Vim~\cite{bimamba} and VMamba~\cite{liu2024vmamba} extend Mamba to 2D images via rasterized or multi-directional scans, but remain largely agnostic to the structural properties of WSIs.  
 
In contrast, our method exploits the intrinsic multi-resolution hierarchy of histopathology images by organizing patch tokens into a region-nested sequence. This aligns SSM state evolution with biologically meaningful coarse-to-fine dependencies, enabling structure-aware WSI modeling without dense self-attention.

\subsection{Mixture-of-Experts Architectures}

Mixture-of-Experts (MoE) architectures increase model capacity through conditional computation, activating only a sparse subset of experts for each input token~\cite{shazeer2017moe,riquelme2021vmoe}.
Recent works explore improved routing mechanisms or domain adaptations, including ExpertFlow~\cite{he2024expertflowoptimizedexpertactivation}, GraphMETRO~\cite{wu2024graphmetro}, and MoVA~\cite{zong2024mova}.
These methods primarily rely on learned gating functions to assign tokens to experts.

However, purely learned routing may be suboptimal in domains such as histopathology, where strong structural priors are available.
We therefore introduce a hybrid expert design that combines deterministic and learned specialization.
Specifically, static resolution experts are deterministically assigned to tokens of different resolutions, enforcing consistent scale-specific encoding.
On top of the resulting region-nested sequence, dynamic sparse experts are activated via learned routing to capture heterogeneous diagnostic patterns across spatial regions.

This dual expert paradigm decouples resolution-aware representation learning from region-adaptive contextual modeling, enabling both structured inductive bias and flexible specialization within a unified framework.

\section{Method}

\subsection{Problem Definition}

We consider whole-slide image (WSI) analysis as a multiple instance learning problem. Given a WSI $\mathcal{W}$, we extract a multi-resolution patch hierarchy: for each resolution level $r \in \{1, \dots, R\}$, we partition $\mathcal{W}$ into patches $\mathcal{P}_r = \{N_r^1, N_r^2, \dots\}$ and encode them into feature vectors $\{x_r^i \in \mathbb{R}^d\}$. The spatial containment relationship between patches at different resolutions defines a hierarchical structure: each coarse patch $N_r^i$ spatially contains a set of finer patches $\mathcal{C}_{r+1}(N_r^i) \subset \mathcal{P}_{r+1}$.

Our goal is to learn a function $f: \{\mathcal{P}_1, \dots, \mathcal{P}_R\} \rightarrow \mathcal{Y}$ that maps the multi-resolution patch collection to a slide-level label $y \in \mathcal{Y}$, typically a diagnostic category. The core innovation lies in decoupling two complementary mechanisms: state-space models capture long-range dependencies along the scanned sequence, while mixture-of-experts enables conditional computation and expert specialization for heterogeneous diagnostic patterns.
The overall structure is shown in Figure~\ref{fig:method}.

\subsection{Region-Nested Selective Scan}
\label{sec:scan}
WSIs exhibit a hierarchical multi-resolution structure, where coarse tissue regions spatially contain finer sub-regions and cellular details.
We formalize this structure as a multi-resolution patch hierarchy and propose a \emph{region-nested selective scan} to linearize it into a 1-D token sequence.

Let $\mathcal{P}_1 = \{N_1^1, \dots, N_1^{n_1}\}$ denote the set of coarse-resolution patches.
For each coarse patch $N_1^i$, we define its child patches at higher resolutions based on spatial containment:
\begin{equation}
\mathcal{C}_2(N_1^i) = \{ N_2^{i,1}, N_2^{i,2}, \dots \},
\end{equation}
and recursively,
\begin{equation}
\mathcal{C}_3(N_2^{i,j}) = \{ N_3^{i,j,1}, \dots, N_3^{i,j,k} \}.
\end{equation}

Region-nested selective scan constructs the token sequence by recursively expanding each coarse patch together with all its descendant patches in a depth-first manner.
Formally, the scanned sequence is defined as
\begin{equation}
\mathcal{S}_{\text{N}} =
\Big[
\mathcal{T}(N_1^1),
\mathcal{T}(N_1^2),
\dots,
\mathcal{T}(N_1^{n_1})
\Big],
\end{equation}
where the expansion operator $\mathcal{T}(\cdot)$ is defined recursively as
\begin{equation}
\mathcal{T}(N_1^i) =
\Big[
N_1^i,\;
\mathcal{T}(N_2^{i,1}),\;
\mathcal{T}(N_2^{i,2}),\;
\dots
\Big],
\end{equation}
and for leaf-level patches,
\begin{equation}
\mathcal{T}(N_3^{i,j,k}) = [\, N_3^{i,j,k} \,].
\end{equation}

This procedure produces a sequence of the form
\begin{align}
\mathcal{S}_{\text{N}} = \big[ 
& N_1^1,\; N_2^{1,1},\; N_3^{1,1,1},\; \dots, \notag \\
& \ \ \ \ \ \ \ \ N_2^{1,2},\; N_3^{1,2,1},\; \dots,  N_1^2,\; \dots \big], \label{eq:Sn}
\end{align}
where all patches belonging to the same coarse region form a contiguous subsequence.

\subsection{Static and Dynamic Experts for Conditional Computation}

To address the inherent heterogeneity of WSIs, we design two complementary expert mechanisms within a single end-to-end framework: (1) static experts for resolution-aware encoding, and (2) dynamic experts for content-adaptive modeling. This dual-expert design enables MoEMambaMIL to exploit both known structural priors from multi-resolution preprocessing and learned content-based specialization from data-driven routing.

\subsubsection{Static Experts for Multi-Resolution Encoding}

Static experts are applied prior to region-nested selective scanning, serving as resolution-specific feature encoders. Each expert is dedicated to a single resolution level and processes only tokens from that scale—a design motivated by the distinct morphological characteristics across magnifications: high-resolution patches capture fine-grained cellular details, while low-resolution ones encode global tissue architecture and spatial context.

\textbf{Architecture}. We implement static experts as a collection of independent Mamba-based encoders, one for each resolution level $r \in \{0, 1, \ldots, R-1\}$. Each encoder consists of $L_{\text{static}}$ stacked Mamba layers with residual connections and layer normalization:

\begin{equation}
\text{MambaStack}_r = \left( \text{LN} \rightarrow \text{Mamba} \rightarrow \text{Residual} \right)^{L_{\text{static}}}
\end{equation}

Formally, given an input token $x_i \in \mathbb{R}^D$ with resolution index $r_i \in \{0, 1, \ldots, R-1\}$, the static expert output is computed as:
\begin{equation}
\tilde{x}_i = f^{\text{static}}_{r_i}(x_i) = \text{MambaStack}_{r_i}(x_i),
\end{equation}
where $f^{\text{static}}_{r}: \mathbb{R}^D \rightarrow \mathbb{R}^D$ denotes the expert specialized for resolution $r$.

\textbf{Hard Assignment Routing}. Unlike dynamic routing mechanisms that learn soft assignments, static experts employ deterministic hard assignment based on the resolution metadata. Given a batch of tokens $\{x_i\}_{i=1}^N$ with corresponding resolution indices $\{r_i\}_{i=1}^N$, we partition the tokens into $R$ disjoint subsets:
\begin{equation}
\mathcal{S}_r = \{i : r_i = r\}, \quad r = 0, 1, \ldots, R-1.
\end{equation}

Each subset $\mathcal{S}_r$ is processed independently by its corresponding expert $f^{\text{static}}_r$. This partitioning ensures that tokens from different resolutions do not interfere during the encoding phase, preserving resolution-specific inductive biases.

\textbf{Implementation Details}. In practice, we implement the resolution-aware encoding efficiently by masking and gathering operations. For each resolution level $r$, we extract the subset of tokens belonging to that resolution, process them through the corresponding Mamba stack, and scatter the outputs back to their original positions:
\begin{equation}
\tilde{X}[\mathcal{S}_r, :] = f^{\text{static}}_r(X[\mathcal{S}_r, :]),
\end{equation}
where $X \in \mathbb{R}^{N \times D}$ denotes the token matrix and $\tilde{X}$ denotes the encoded output.

By explicitly specializing experts to resolution-specific representations, static experts exploit the strong physical prior encoded by multi-resolution WSI preprocessing. This design enables resolution-aware feature encoding while avoiding unnecessary routing complexity at this early stage. The deterministic assignment also ensures stable gradient flow during training, as there is no discrete routing decision to optimize.

\subsubsection{Dynamic Experts for Region-Aware Modeling}

After region-nested selective scanning, the tokens are reorganized into a region-aware sequence that captures both local morphological patterns and global spatial relationships. On top of this scanned sequence, we employ dynamic experts implemented via a sparse mixture-of-experts (SparseMoE) mechanism integrated with Mamba layers.

\textbf{Architecture}. The dynamic expert module consists of $L_{\text{dyn}}$ MoE-Mamba blocks, each containing two sub-layers: (1) a standard Mamba layer for sequential modeling, and (2) a SparseMoE layer for conditional computation. Both sub-layers employ pre-normalization and residual connections:
\begin{equation}
h^{(l)} = h^{(l-1)} + \text{Mamba}\left(\text{LN}(h^{(l-1)})\right),
\end{equation}
\begin{equation}
h^{(l+1)} = h^{(l)} + \text{SparseMoE}\left(\text{LN}(h^{(l)})\right).
\end{equation}

\textbf{Gating Mechanism}. Unlike static experts, dynamic experts are shared across all regions and resolutions. A lightweight gating network performs token-level routing, assigning each token to a small subset of $k$ experts (out of $E$ total experts) based on its content. The gating network is implemented as a single linear projection:
\begin{equation}
g_i = W_g \tilde{x}_i + b_g \in \mathbb{R}^E,
\end{equation}
where $W_g \in \mathbb{R}^{E \times D}$ and $b_g \in \mathbb{R}^E$ are learnable parameters.

\textbf{Top-K Sparse Routing}. To enable sparse computation, we select only the top-$k$ experts with the highest gating scores for each token:
\begin{equation}
\mathcal{E}_i = \text{TopK}(g_i, k) = \{e_1, e_2, \ldots, e_k\},
\end{equation}
where $\mathcal{E}_i$ denotes the set of selected expert indices for token $i$. The routing weights are computed by applying softmax over the selected experts:
\begin{equation}
\alpha_{i,e} = \frac{\exp(g_{i,e})}{\sum_{e' \in \mathcal{E}_i} \exp(g_{i,e'})}, \quad e \in \mathcal{E}_i.
\end{equation}

Note that when $k=1$, the routing weight degenerates to $\alpha_{i,e} = 1$ for the single selected expert, reducing to hard routing.

\textbf{Expert Architecture}. Each dynamic expert $f^{\text{dyn}}_e$ is implemented as a Mamba-based module with layer normalization and residual connection:
\begin{equation}
f^{\text{dyn}}_e(x) = x + \text{Mamba}_e(\text{LN}_e(x)),
\end{equation}
where $\text{Mamba}_e$ and $\text{LN}_e$ denote the expert-specific Mamba layer and layer normalization, respectively. This design allows each expert to specialize in capturing different sequential patterns while maintaining the state-space modeling capability.

\textbf{Sparse Computation}. The final output for each token is computed as a weighted combination of the selected expert outputs:
\begin{equation}
y_i = \sum_{e \in \mathcal{E}_i} \alpha_{i,e} \cdot f^{\text{dyn}}_e(\tilde{x}_i).
\end{equation}

To achieve true sparse computation, we implement expert-wise token routing rather than token-wise expert evaluation. Specifically, for each expert $e$, we gather all tokens routed to it:
\begin{equation}
\mathcal{T}_e = \{i : e \in \mathcal{E}_i\},
\end{equation}
process them in a single batched forward pass through $f^{\text{dyn}}_e$, and scatter the weighted outputs back to the corresponding positions. This implementation avoids redundant computation and enables efficient GPU utilization.

\subsubsection{Load Balancing Regularization}

A well-known challenge in training MoE models is expert collapse, where the gating network converges to routing all tokens to a small subset of experts, leaving other experts underutilized. To prevent this degenerate behavior, we adopt a standard load-balancing regularization that encourages uniform expert utilization across tokens.

For each expert $e$, we define two metrics:
\textbf{Importance $\bar{p}_e$:} the average routing probability assigned to expert $e$ across all tokens:
\begin{equation}
\bar{p}_e = \frac{1}{BN} \sum_{b=1}^{B} \sum_{i=1}^{N} p_{b,i,e}, \quad p_{b,i,e} = \text{softmax}(g_{b,i})_e.
\end{equation}
\textbf{Load $\bar{c}_e$:} the fraction of tokens for which expert $e$ is the top-1 choice:
\begin{equation}
\bar{c}_e = \frac{1}{BN} \sum_{b=1}^{B} \sum_{i=1}^{N} \mathbb{1}[\arg\max_e g_{b,i} = e].
\end{equation}

\textbf{Auxiliary Loss}. The load-balancing loss is defined as the scaled dot product of importance and load vectors:
\begin{equation}
\mathcal{L}_{\text{balance}} = E \sum_{e=1}^{E} \bar{p}_e \cdot \bar{c}_e.
\end{equation}

This formulation penalizes configurations where both importance and load are concentrated on the same experts. The scaling factor $E$ ensures that the loss magnitude is independent of the number of experts. 
The final load-balancing loss becomes:
\begin{equation}
\mathcal{L}_{\text{balance}} = \frac{1}{L_{\text{dyn}}} \sum_{l=1}^{L_{\text{dyn}}} \mathcal{L}^{(l)}_{\text{balance}}.
\end{equation}

\subsection{Training Objective}

\textbf{Slide-Level Aggregation}. Following the multiple instance learning paradigm, we aggregate token-level representations into a slide-level representation using attention-based pooling:
\begin{equation}
a_i = \frac{\exp(w^\top \tanh(V h_i))}{\sum_{j=1}^{N} \exp(w^\top \tanh(V h_j))},
\end{equation}
\begin{equation}
z = \sum_{i=1}^{N} a_i h_i,
\end{equation}
where $V \in \mathbb{R}^{D' \times D}$ and $w \in \mathbb{R}^{D'}$ are learnable parameters, and $z \in \mathbb{R}^D$ is the aggregated slide representation. The final prediction is obtained via a linear classifier:
\begin{equation}
\hat{y} = \text{softmax}(W_{\text{cls}} z + b_{\text{cls}}).
\end{equation}

The overall training objective consists of two components:
\begin{equation}
\mathcal{L} = \mathcal{L}_{\text{task}} + \lambda \mathcal{L}_{\text{balance}},
\end{equation}
where $\mathcal{L}_{\text{task}}$ is the task-specific loss and $\mathcal{L}_{\text{balance}}$ is the load-balancing regularization defined above.

\section{Experiments}

\subsection{Dataset and Implementations}

Our study utilizes a total of 2,355 WSIs from three multi-class datasets. The TCGA kidney dataset consists of 887 WSIs covering three renal carcinoma subtypes, including 500 KIRC, 273 KIRP, and 114 KICH cases. The liver cancer dataset contains 968 WSIs, comprising 670 HCC, 206 CHC, and 92 ICC cases. The CAMELYON17 dataset includes 500 breast WSIs (five slides per patient) with metastasis annotations, including 87 macro-metastases, 95 micro-metastases (including ITCs), and 318 normal slides.

For fair comparison, all methods follow the preprocessing pipeline of CLAM~\cite{lu2021clam} and the experimental protocol of CP-MID~\cite{wu26tip}. WSIs are processed at $5\times$, $10\times$, and $20\times$ magnifications, with non-overlapping $256\times256$ patches extracted using OpenSlide. Tissue regions are identified by Otsu thresholding on the HSV saturation channel, and patches with over 30\% background are removed.  
All models are trained using Adam with a learning rate of $1\times10^{-4}$, a batch size of 1, and up to 15 epochs. Datasets are split at the patient level following CP-MID, with fixed splits shared across all methods to ensure reproducibility and fair evaluation.

\subsection{Performance Comparison with Existing Works}

We compare our approach with several state-of-the-art methods, including DSMIL \cite{li2021dual}, TransMIL\cite{shao2021transmil}, CLAM\cite{lu2021clam}, UNI\cite{chen2024towards}, Prov-GigaPath\cite{xu2024whole}, as well as MambaMIL\cite{gu2024mamba}, BiMambaMIL\cite{bimamba}, and SRMambaMIL\cite{mambamil}.

As shown in Table~\ref{tab:sota}, MoEMambaMIL consistently outperforms existing MIL methods across most evaluation settings on TCGA Kidney, Liver Cancer, and Camelyon17 datasets. Under all three feature extractors (ResNet\cite{resnet}, UNI\cite{chen2024towards}, and GigaPath\cite{xu2024whole}), MoEMambaMIL achieves the highest or near-highest F1, AUC, and accuracy. In particular, it achieves state-of-the-art performance on TCGA Kidney (F1 of 95.78\% with UNI features) and shows advantages on the challenging Camelyon17 dataset (up to 89.99\% F1 with Giga features), indicating superior generalization under domain shifts. Compared to both attention-based MIL methods and recent Mamba-based variants, the consistent improvements highlight the effectiveness of the proposed mixture-of-experts Mamba architecture. Moreover, while stronger feature extractors (UNI and Giga) generally improve all methods, MoEMambaMIL benefits the most, confirming its ability to leverage high-quality representations for whole-slide image classification.

\begin{table*}[]
\small
\centering
\caption{Performance comparison on three datasets with three feature extractors. RN: ResNet. Giga: GigaPath. $^\dagger$: from \cite{wu26tip}. } % Prov-GigaPath(LP) denotes linear probing on frozen pretrained weights. Other baselines are fully fine-tuned by default. 
\begin{tabular}{c|c|ccc|ccc|ccc}
\hline
              &       & \multicolumn{3}{c|}{TCGA Kidney} & \multicolumn{3}{c|}{Liver Cancer} & \multicolumn{3}{c}{Camelyon 17} \\ \hline
Methods       & Feat. & \multicolumn{1}{c|}{F1} & \multicolumn{1}{c|}{AUC} & ACC & \multicolumn{1}{c|}{F1} & \multicolumn{1}{c|}{AUC} & ACC & \multicolumn{1}{c|}{F1} & \multicolumn{1}{c|}{AUC} & ACC \\ \hline
DSMIL$^\dagger$         & RN    & \multicolumn{1}{c|}{66.24}  & \multicolumn{1}{c|}{95.47}    & 80.00           & \multicolumn{1}{c|}{87.42}  & \multicolumn{1}{c|}{98.20}  & 92.64           & \multicolumn{1}{c|}{54.50}           & \multicolumn{1}{c|}{77.10}    & 76.00           \\
TransMIL$^\dagger$      & RN    & \multicolumn{1}{c|}{73.78}  & \multicolumn{1}{c|}{95.27}    & 84.04           & \multicolumn{1}{c|}{70.24}  & \multicolumn{1}{c|}{94.32}  & 84.56           & \multicolumn{1}{c|}{65.72}           & \multicolumn{1}{c|}{82.36}    & 86.20           \\
CLAM$^\dagger$          & RN    & \multicolumn{1}{c|}{64.75}  & \multicolumn{1}{c|}{94.84}    & 79.55           & \multicolumn{1}{c|}{84.24}  & \multicolumn{1}{c|}{98.43}  & 92.12           & \multicolumn{1}{c|}{63.52}           & \multicolumn{1}{c|}{77.03}    & 87.00           \\  
CP-MID$^\dagger$        & RN    & \multicolumn{1}{c|}{84.21}  & \multicolumn{1}{c|}{97.47}    & 88.88           & \multicolumn{1}{c|}{86.65}  & \multicolumn{1}{c|}{97.98}  & 92.46           & \multicolumn{1}{c|}{62.66}           & \multicolumn{1}{c|}{78.37}    & 86.40           \\
CLAM$^\dagger$          & UNI   & \multicolumn{1}{c|}{90.53}  & \multicolumn{1}{c|}{99.14}    & 92.13           & \multicolumn{1}{c|}{83.72}  & \multicolumn{1}{c|}{98.36}  & 92.96           & \multicolumn{1}{c|}{73.97}           & \multicolumn{1}{c|}{85.57}    & 87.00           \\
CLAM$^\dagger$          & Giga  & \multicolumn{1}{c|}{92.20}  & \multicolumn{1}{c|}{99.21}    & 93.82           & \multicolumn{1}{c|}{84.27}  & \multicolumn{1}{c|}{98.02}  & 93.07           & \multicolumn{1}{c|}{78.57}           & \multicolumn{1}{c|}{90.90}    & 88.60           \\
Prov-GigaPath$^\dagger$ & Giga  & \multicolumn{1}{c|}{81.52}  & \multicolumn{1}{c|}{95.74}    & 84.83           & \multicolumn{1}{c|}{69.39}  & \multicolumn{1}{c|}{94.21}  & 82.51           & \multicolumn{1}{c|}{63.20}           & \multicolumn{1}{c|}{53.20}    & 41.46           \\
CP-MID$^\dagger$        & Giga  & \multicolumn{1}{c|}{93.61}  & \multicolumn{1}{c|}{99.41}    & 95.28           & \multicolumn{1}{c|}{\textbf{88.95}}               & \multicolumn{1}{c|}{98.78}  & 93.78           & \multicolumn{1}{c|}{81.75}           & \multicolumn{1}{c|}{89.68}    & 93.00           \\ \hline
MambaMIL      & RN    & \multicolumn{1}{c|}{81.16}  & \multicolumn{1}{c|}{98.08}    & 85.39           & \multicolumn{1}{c|}{84.03}  & \multicolumn{1}{c|}{97.04}  & 90.78           & \multicolumn{1}{c|}{62.72}           & \multicolumn{1}{c|}{74.90}    & 91.00           \\
BiMambaMIL    & RN    & \multicolumn{1}{c|}{82.00}  & \multicolumn{1}{c|}{98.19}    & 88.20           & \multicolumn{1}{c|}{86.22}  & \multicolumn{1}{c|}{97.98}  & 91.18           & \multicolumn{1}{c|}{62.68}           & \multicolumn{1}{c|}{82.06}    & 92.00           \\
SRMambaMIL    & RN    & \multicolumn{1}{c|}{82.97}  & \multicolumn{1}{c|}{97.63}    & 87.64           & \multicolumn{1}{c|}{86.98}  & \multicolumn{1}{c|}{97.76}  & 91.57           & \multicolumn{1}{c|}{69.93}           & \multicolumn{1}{c|}{80.86}    & 92.00           \\
MoEMambaMIL   & RN    & \multicolumn{1}{c|}{{ 88.71}} & \multicolumn{1}{c|}{98.19}   & 91.57          & \multicolumn{1}{c|}{87.29}  & \multicolumn{1}{c|}{\textbf{98.91}}               & 93.47           & \multicolumn{1}{c|}{70.52}          & \multicolumn{1}{c|}{83.37}   & 92.00          \\ \hline
MambaMIL      & UNI   & \multicolumn{1}{c|}{93.64} & \multicolumn{1}{c|}{98.56}   & 94.94          & \multicolumn{1}{c|}{76.20}  & \multicolumn{1}{c|}{97.67}  & 91.96           & \multicolumn{1}{c|}{76.99}          & \multicolumn{1}{c|}{91.40}   & 92.00          \\
BiMambaMIL    & UNI   & \multicolumn{1}{c|}{94.16} & \multicolumn{1}{c|}{99.41}   & 94.94          & \multicolumn{1}{c|}{84.94}  & \multicolumn{1}{c|}{97.04}  & 93.97           & \multicolumn{1}{c|}{75.21}          & \multicolumn{1}{c|}{91.25}   & 93.00          \\
SRMambaMIL    & UNI   & \multicolumn{1}{c|}{{ 92.84}} & \multicolumn{1}{c|}{99.21}   & 93.82          & \multicolumn{1}{c|}{71.82}  & \multicolumn{1}{c|}{96.93}  & 91.46           & \multicolumn{1}{c|}{{ 81.38}}          & \multicolumn{1}{c|}{94.18}   & 93.00          \\
MoEMambaMIL   & UNI   & \multicolumn{1}{c|}{\textbf{95.78}}               & \multicolumn{1}{c|}{{ \textbf{99.44}}}                & { \textbf{96.63}} & \multicolumn{1}{c|}{87.42}  & \multicolumn{1}{c|}{98.36}  & 94.47           & \multicolumn{1}{c|}{88.81}          & \multicolumn{1}{c|}{97.54}   & 96.00          \\ \hline
MambaMIL      & Giga  & \multicolumn{1}{c|}{93.77} & \multicolumn{1}{c|}{99.04}   & 94.94          & \multicolumn{1}{c|}{81.83} & \multicolumn{1}{c|}{97.26} & 91.96          & \multicolumn{1}{c|}{79.82}          & \multicolumn{1}{c|}{92.82} & 94.00          \\
BiMambaMIL    & Giga  & \multicolumn{1}{c|}{89.96} & \multicolumn{1}{c|}{99.32}   & 91.57          & \multicolumn{1}{c|}{84.13} & \multicolumn{1}{c|}{97.82} & 93.47          & \multicolumn{1}{c|}{80.49}          & \multicolumn{1}{c|}{93.53}   & 94.00          \\
SRMambaMIL    & Giga  & \multicolumn{1}{c|}{90.73} & \multicolumn{1}{c|}{98.35} & 93.82          & \multicolumn{1}{c|}{78.81} & \multicolumn{1}{c|}{97.49} & 91.46          & \multicolumn{1}{c|}{79.82}          & \multicolumn{1}{c|}{90.53}   &  94.00          \\
MoEMambaMIL   & Giga  & \multicolumn{1}{c|}{95.16} & \multicolumn{1}{c|}{99.43}   & { \textbf{96.63}} & \multicolumn{1}{c|}{85.59} & \multicolumn{1}{c|}{98.25} & { \textbf{94.47}} & \multicolumn{1}{c|}{{ \textbf{89.99}}} & \multicolumn{1}{c|}{{ \textbf{95.24}}}                & { \textbf{96.00}} \\ \hline
\end{tabular}
\label{tab:sota}
\end{table*}

\subsection{Ablation Study}

\subsubsection{Complementarity of Selective Scan}
\label{sec:scan}

\begin{figure}
    \centering
    \includegraphics[width=1\linewidth]{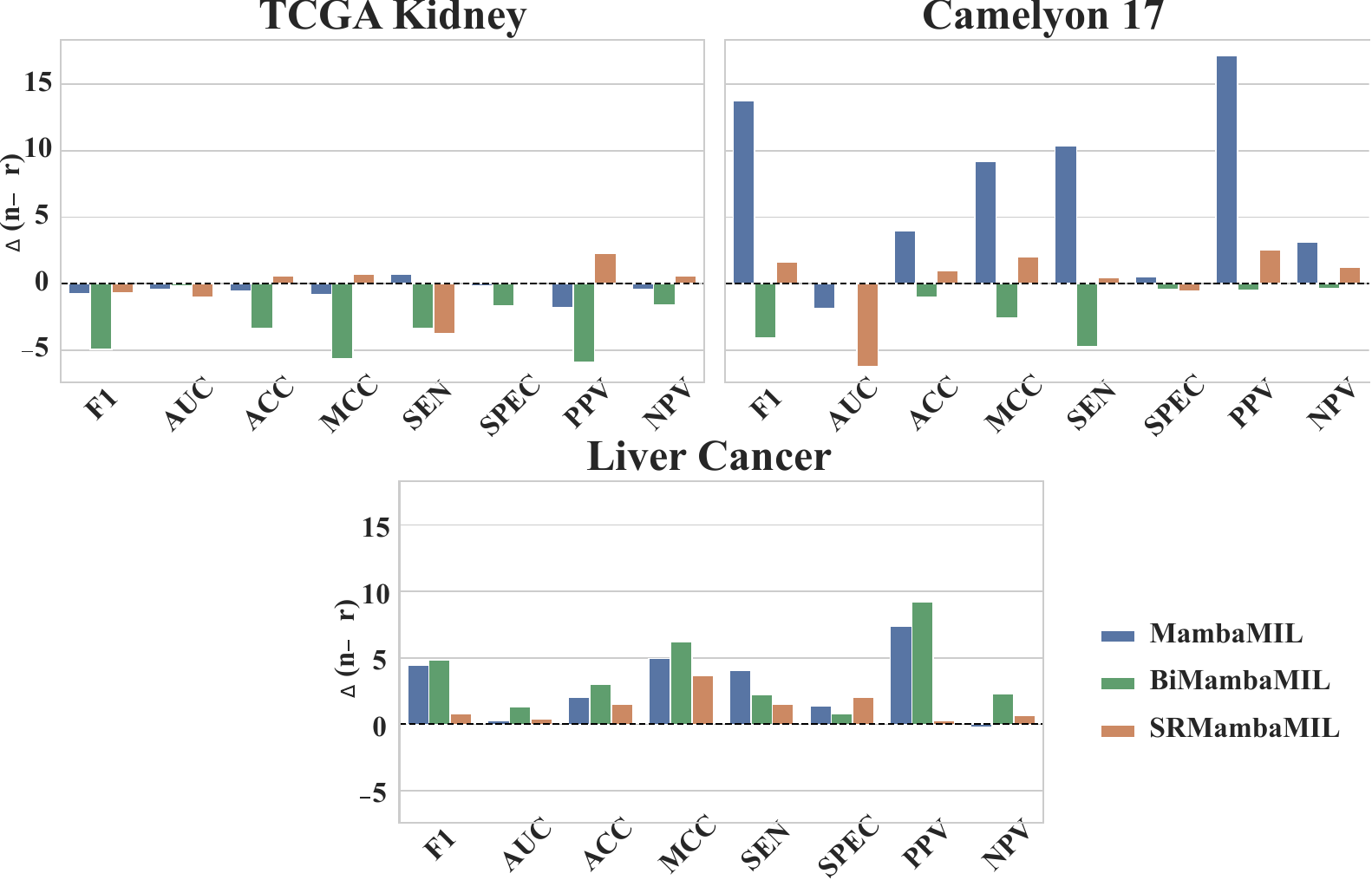}
    \caption{
    Performance differences ($\Delta = n - r$) between resolution-based (r) and region-nested (n) selective scanning across datasets and models. The mixed signs across metrics indicate complementary strengths of the two schemes.
    }
    \label{fig:scan}
\end{figure}

As shown in Figure~\ref{fig:scan}, neither the resolution-based nor the region-nested selective scan consistently outperforms the other across all datasets, models, and evaluation metrics. In several cases, resolution-based scanning yields higher specificity or AUC, while region-nested scanning demonstrates advantages in sensitivity or F1-score.

This observation suggests that the two scanning schemes emphasize different aspects of the feature space.
Resolution-based selective scan tends to prioritize global structural patterns across multiple resolutions, whereas region-nested selective scan focuses on localized hierarchical relationships within anatomically coherent regions.

By allowing Mamba-based MIL to access both global resolution-aware cues and region-level nested representations, our approach is better positioned to capture diverse discriminative signals. Figure~\ref{fig:scan} indicates that resolution and region nested-selective scan schemes are not competing alternatives but complementary components, whose integration is crucial for robust representation learning.

\subsubsection{Component-wise Ablation}

Our proposed MoEMambaMIL employs a resolution-aware static expert to model resolution-aware sequences, while leveraging a dynamic Mamba-based Mixture-of-Experts (MoE) architecture to capture region-nested sequence dependencies.

\begin{figure}
    \centering
    \includegraphics[width=0.8\linewidth]{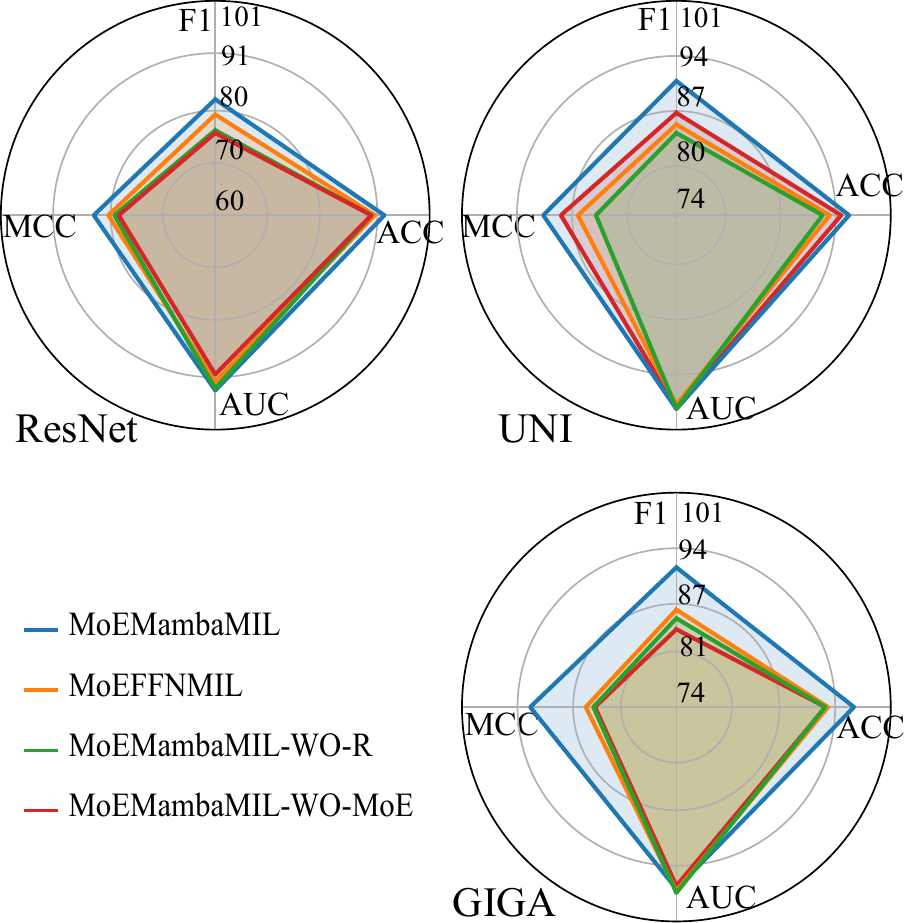}
    \caption{Ablation study across different encoders (ResNet, UNI, Gigapath): Mean Performance Across Metrics. }
    \label{fig:abla-radar}
\end{figure}

To systematically evaluate the contribution of each component, we design several ablation variants: (1). WO-R: removes the resolution-aware sequence modeling, retaining only region-level nested sequence modeling. (2). WO-MoE: replaces the MoE module with a single expert, disabling dynamic expert selection. (3). MoEFFNMIL: replaces the Mamba-based MoE with a feed-forward-network (FFN)-based MoE while keeping the same expert routing strategy.
The quantitative results are reported in Table~\ref{tab:abla}.
In addition, we average the performance across different encoders (ResNet, UNI, and Gigapath) and visualize the results in Figure~\ref{fig:abla-radar}.

Our ablation study validates the effectiveness of each core component in MoEMambaMIL (Table \ref{tab:abla}). First, the resolution-aware expert is crucial; removing it (WO/R) consistently degrades performance, most notably on Liver Cancer where F1 drops by ~7\%, confirming the importance of multi-scale spatial modeling in pathology images. Second, the dynamic MoE structure provides substantial gains; the single-expert variant (WO/MoE) performs worst overall, with a dramatic ~10\% F1 decrease on Liver Cancer, demonstrating that adaptive expert routing captures diverse tissue patterns more effectively. Third, the Mamba-based sequence modeling outperforms a standard FFN-based MoE, verifying Mamba’s superior capability in modeling long-range, region-nested dependencies. Collectively, these results affirm that combining resolution-aware modeling with a Mamba-driven dynamic MoE yields a robust architecture for whole-slide image analysis.

\begin{table}[]
\small
\centering
\caption{Ablation on each component. Dataset: Camelyon 17. Patch-level Encoder: Prov-Gigapath. }
\label{tab:abla}
\begin{tabular}{c|ccc}
\hline
\multicolumn{4}{c}{\textbf{TCGA Kidney}} \\ \hline
Methods     & F1    & AUC    & ACC    \\ \hline
WO/MoE      & 93.11 & 99.11 & 94.38 \\
WO/R        & 91.62 & 99.08 & 93.26 \\
MoEFFNMIL      & 92.88 & 99.15 & 93.82 \\
MoEMambaMIL & \textbf{95.16} & \textbf{99.43} & \textbf{96.63} \\ \hline
\end{tabular}
\vspace{0.4cm}
\begin{tabular}{c|ccc}
\hline
\multicolumn{4}{c}{\textbf{Liver Cancer}} \\ \hline
Methods     & F1    & AUC    & ACC    \\ \hline
WO/MoE      & 75.39 & 97.11 & 89.95 \\
WO/R        & 78.52 & 97.64 & 90.95 \\
MoEFFNMIL      & 75.92 & 97.09 & 89.45 \\
MoEMambaMIL & \textbf{85.59} & \textbf{98.25} & \textbf{94.47} \\ \hline
\end{tabular}
\vspace{0.4cm}
\begin{tabular}{c|ccc}
\hline
\multicolumn{4}{c}{\textbf{Camelyon 17}} \\ \hline
Methods     & F1    & AUC    & ACC    \\ \hline
WO/MoE      & 82.88 & 93.37 & 94.00 \\
WO/R        & 85.39 & 95.48 & 94.00 \\
MoEFFNMIL      & 87.01 & 92.95 & 95.00 \\
MoEMambaMIL & \textbf{89.99} & \textbf{95.24} & \textbf{96.00} \\ \hline
\end{tabular}
\end{table}

\subsubsection{Effect of MoE-Mamba Design}

We analyze key design choices of the proposed MoE-Mamba architecture, including network depth, expert activation sparsity, and the load balancing loss weight $\mathcal{L}_{\text{balance}}$.
MoE modules are inserted into every Mamba layer. All results are reported using macro-averaged F1-score, AUC, ACC and MCC for multi-class evaluation.

As shown in Table~\ref{tab:analysis}, increasing the number of MoE-Mamba layers consistently improves performance, with gains saturating at six layers.
We further observe that activating two experts per token yields the best overall performance, achieving a favorable balance between expert specialization and robustness.
Regarding routing regularization, a small load balancing loss weight ($\lambda=0.001$) results in stable and superior performance, while larger values degrade macro-F1 and macro-MCC. Based on these results, we adopt a six-layer MoE-Mamba architecture with four experts per layer, two activated experts, and $\lambda=0.001$ for all experiments unless otherwise specified.

\begin{table}[]
\small
\caption{Ablation Study of MoE-Mamba Architecture Design.}
\begin{tabular}{ccccccc} \hline
Layers & TopK & $\mathcal{L}_{\text{balance}}$ & F1    & AUC   & ACC   & MCC   \\ \hline
2      & 2                 & 0.001             & 82.98 & 91.52 & 95.00 & 86.76 \\
3      & 2                 & 0.001             & 83.16 & 92.50 & 95.00 & 86.85 \\
4      & 2                 & 0.001             & 87.00 & 91.70 & 95.00 & 87.19 \\
5      & 2                 & 0.001             & 88.86 & 92.11 & 96.00 & 89.15 \\
6      & 2                 & 0.001             & 89.99 & 95.24 & 96.00 & 90.00 \\
6      & 1                 & 0.001             & 84.57 & 97.76 & 95.00 & 86.94 \\
6      & 2                 & 0.001             & 89.99 & 95.24 & 96.00 & 90.00 \\
6      & 3                 & 0.001             & 86.15 & 93.45 & 95.00 & 87.05 \\
6      & 4                 & 0.001             & 90.86 & 92.39 & 97.00 & 92.15 \\
6      & 2                 & 0.001             & 89.99 & 95.24 & 96.00 & 90.00 \\
6      & 2                 & 0.01              & 90.86 & 94.18 & 97.00 & 92.15 \\
6      & 2                 & 0.1               & 87.23 & 92.52 & 96.00 & 89.46 \\
6      & 2                 & 1                 & 90.86 & 93.33 & 97.00 & 92.15 \\  \hline
\end{tabular}
\label{tab:analysis}
\end{table}

\subsection{Qualitative Results}

\begin{figure}
    \centering
    \includegraphics[width=1\linewidth]{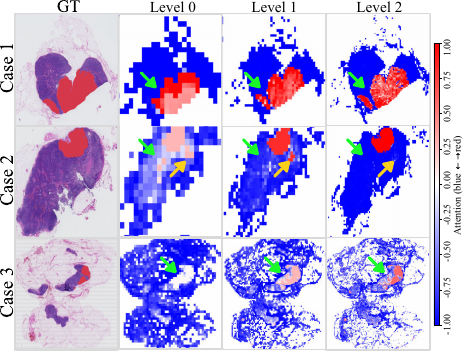}
    \caption{Multi-resolution MIL attention visualization.
    For each WSI, visualizing instance-level attention weights derived from the final MIL pooling for three resolution levels (Level 0–2). Red indicates high attention, blue indicates low attention, and white denotes intermediate importance.}
    \label{fig:heatmap}
\end{figure}

Figure~\ref{fig:heatmap} shows that MoEMambaMIL consistently attends to ground-truth regions across all resolutions, while different resolutions exhibit distinct but complementary behaviors. Coarser resolutions provide robust global localization but may blur fine structures or include false positives, whereas finer resolutions progressively sharpen attention and suppress ambiguous regions. This resolution-dependent specialization enables accurate and robust slide-level prediction by aggregating complementary evidence across scales.
We provide a detailed qualitative analysis in the Appendix~\ref{appendix:Qualitative}.

\section{Conclusion}

In this paper, we propose MoEMambaMIL, a multi-resolution multiple instance learning framework for whole-slide image analysis that combines region-nested selective scanning with a Mamba-based mixture-of-experts architecture. By decoupling resolution-aware static experts from content-adaptive dynamic experts, the proposed method effectively captures both multi-scale structural priors and heterogeneous regional patterns in WSIs. Extensive experiments on three public pathology benchmarks demonstrate consistent and robust improvements over state-of-the-art MIL and Mamba-based methods across different feature extractors. These results highlight the effectiveness of integrating state-space modeling and conditional computation for large-scale histopathology analysis.

\section{Limitations and Future Work}

First, the proposed region-nested selective scan follows a predefined hierarchy derived from multi-resolution preprocessing and does not learn the scan order end-to-end, which may restrict adaptability to irregular spatial structures. Second, although sparse routing reduces computation, the Mixture-of-Experts design increases model and training complexity, and careful configuration of expert number and routing sparsity is required for scalability. Finally, our experiments focus on slide-level classification; the applicability of MoEMambaMIL to other weakly supervised or structured prediction tasks remains to be explored.

\nocite{langley00}

\bibliography{main}
\bibliographystyle{icml2026}

%%%%%%%%%%%%%%%%%%%%%%%%%%%%%%%%%%%%%%%%%%%%%%%%%%%%%%%%%%%%%%%%%%%%%%%%%%%%%%%
%%%%%%%%%%%%%%%%%%%%%%%%%%%%%%%%%%%%%%%%%%%%%%%%%%%%%%%%%%%%%%%%%%%%%%%%%%%%%%%
% APPENDIX
%%%%%%%%%%%%%%%%%%%%%%%%%%%%%%%%%%%%%%%%%%%%%%%%%%%%%%%%%%%%%%%%%%%%%%%%%%%%%%%
%%%%%%%%%%%%%%%%%%%%%%%%%%%%%%%%%%%%%%%%%%%%%%%%%%%%%%%%%%%%%%%%%%%%%%%%%%%%%%%
\newpage
\appendix
\onecolumn
\section{Appendix.}

% You can have as much text here as you want. The main body must be at most $8$
% pages long. For the final version, one more page can be added. If you want, you
% can use an appendix like this one.

% The $\mathtt{\backslash onecolumn}$ command above can be kept in place if you
% prefer a one-column appendix, or can be removed if you prefer a two-column
% appendix.  Apart from this possible change, the style (font size, spacing,
% margins, page numbering, etc.) should be kept the same as the main body.

\subsection{Algorithm}

\begin{algorithm}[t]
\caption{Resolution-Aware Static Encoding}
\label{alg:static}
\begin{algorithmic}[1]

\STATE \textbf{Input:}
Patch features $X = \{x_i\}_{i=1}^N$,
resolution indices $\{r_i\}_{i=1}^N$

\STATE \textbf{Parameters:}
Resolution-specific encoders
$\{f^{\text{static}}_r\}_{r=1}^R$

\STATE \textbf{Output:}
Encoded features $\tilde{X} = \{\tilde{x}_i\}_{i=1}^N$

\FOR{$r = 1$ to $R$}
    \STATE $\mathcal{I}_r \gets \{ i \mid r_i = r \}$
    \STATE $\tilde{X}[\mathcal{I}_r] \gets
    f^{\text{static}}_r(X[\mathcal{I}_r])$
\ENDFOR

\STATE \textbf{Return} $\tilde{X}$

\end{algorithmic}
\end{algorithm}

\begin{algorithm}[t]
\caption{MoEMamba Block on Region-Nested Scanned Sequences}
\label{alg:moemamba}
\begin{algorithmic}[1]

\STATE \textbf{Input:} 
Region-nested scanned token sequence
$X = (x_1, \dots, x_L)$ (encoded by Algorithm.A~\ref{alg:static})

\STATE \textbf{Parameters:} 
Number of experts $E$, top-$k$ routing parameter $k$

\STATE \textbf{Output:} 
Updated token sequence $X' = (x'_1, \dots, x'_L)$

\vspace{0.4em}
\STATE \textbf{Region-nested selective scan (structure definition):}  
$X$ is constructed by the proposed region-nested selective scan
(Sec~\ref{sec:scan}), which induces a hierarchical, locality-preserving order over
multi-resolution tokens.

\vspace{0.4em}
\STATE \textbf{Shared state-space modeling:}  
$\tilde{X} \gets X + \mathrm{SSM}\big(\mathrm{LN}(X)\big)$

\vspace{0.4em}
\STATE \textbf{Token-level sparse routing:}  
$G \gets \mathrm{Gate}\big(\mathrm{LN}(\tilde{X})\big) \in \mathbb{R}^{L \times E}$

\FOR{$i = 1$ to $L$}
    \STATE Select top-$k$ experts:  
    $\mathcal{E}_i \gets \mathrm{TopK}(G_i, k)$
    \STATE Compute routing weights:  
    $\alpha_{i,e} \gets
    \mathrm{Softmax}\big(\{G_{i,e'}\}_{e' \in \mathcal{E}_i}\big),
    \quad e \in \mathcal{E}_i$
\ENDFOR

\vspace{0.4em}
\STATE \textbf{Expert-wise sparse state-space computation:}

\FOR{$e = 1$ to $E$}
    \STATE Collect routed tokens with preserved scan order:  
    $\mathcal{I}_e \gets \{\, i \mid e \in \mathcal{E}_i \,\}$
    \STATE $X^{(e)} \gets (\tilde{x}_i)_{i \in \mathcal{I}_e}$
    \STATE Apply expert-specific SSM:  
    $Y^{(e)} \gets \mathrm{SSM}_e(X^{(e)})$
\ENDFOR

\vspace{0.4em}
\STATE \textbf{Aggregation and residual update:}

\FOR{$i = 1$ to $L$}
    \STATE $x'_i \gets \tilde{x}_i$
    \FOR{each $e \in \mathcal{E}_i$}
        \STATE $x'_i \gets x'_i
        + \alpha_{i,e} \cdot Y^{(e)}_{\pi_e(i)}$
    \ENDFOR
\ENDFOR

\STATE \textbf{Return} $X'$

\end{algorithmic}
\end{algorithm}

\subsection{Evaluation Metrics and Environment}
Model performance is evaluated using standard WSI classification metrics following \cite{wu26tip}, including F1-score, AUC, accuracy (ACC), Matthews correlation coefficient (MCC), sensitivity (SENS), specificity (SPEC), positive predictive value (PPV), and negative predictive value (NPV).

All experiments are performed on a workstation running Ubuntu 18.04.6 LTS, equipped with an NVIDIA V100 GPU (32GB), using CUDA 11.7, Python 3.7.7, and PyTorch 1.12.1+cu113.

\subsection{Notation and Code Correspondence}
\label{appendix:notation}

To facilitate reproducibility and clarity, we provide a detailed mapping between the mathematical notation used in the main text and the corresponding expressions in our implementation (see Table.A~\ref{tab:notation_tensors}), as well as all critical architectural hyperparameters of our model (see Table.A~\ref{tab:hyperparams}). 

\begin{table}[t]
% \small
\centering
\caption{Symbol Mapping for Tensors and Functions}
\label{tab:notation_tensors}
% \resizebox{\textwidth}{!}{%
\begin{tabular}{c|l|c}
\toprule
\textbf{Symbol} & \textbf{Description} & \textbf{Shape} \\
\midrule
$X$ & Raw patch feature matrix from WSI & $[B, N, D_{\text{in}}]$ \\
$r_i$ & Resolution index of each patch & $[B, N]$ \\
$H$ & Patch features after embedding & $[B, N, D]$ \\
$\tilde{H}$ & Features after static expert encoding & $[B, N, D]$ \\
$\mathcal{S}_r$ & Index set of patches at resolution $r$ & - \\
$f^{\text{static}}_r$ & Static expert for resolution $r$ & - \\
$f^{\text{dyn}}_e$ & Dynamic expert with index $e$ & - \\
$g_i$ & Gating scores over all experts & $[B, N, E]$ \\
$\mathcal{E}_i$ & Top-$k$ selected expert indices & $[B, N, k]$ \\
$\alpha_{i,e}$ & Normalized routing weight & $[B, N, k]$ \\
$a_i$ & Attention weight for aggregation & $[B, 1, N]$ \\
$z$ & Slide-level representation & $[B, D]$ \\
$\hat{y}$ & Predicted class logits & $[B, C]$ \\
\bottomrule
\end{tabular}%
\end{table}

\begin{table}[t]
\centering
\caption{Hyperparameter Definitions and Default Settings}
\label{tab:hyperparams}
% \resizebox{\textwidth}{!}{%
\begin{tabular}{l|c|c|l}
\toprule
\textbf{Hyperparameter} & \textbf{Symbol} & \textbf{Default} & \textbf{Description} \\
\midrule
Batch size & $B$ & 1 & Number of WSIs per batch \\
Number of patches & $N$ & varies & Patches per slide \\
Input feature dim & $D_{\text{in}}$ & ResNet: 1024, UNI: 1024, Giga: 1536& Pretrained feature dimension \\
Hidden dimension & $D$ & 512 & Model hidden dimension \\
Number of classes & $C$ & task-specific & Classification categories \\
State dimension & $d_{\text{state}}$ & 16 & Mamba state dimension \\
Convolution width & $d_{\text{conv}}$ & 4 & Mamba convolution kernel \\
Expansion factor & - & 2 & Mamba expansion ratio \\
FFN hidden dim & $d_{\text{hidden}}$ & 1024 & MoE-FFN hidden dimension \\
Number of experts & $E$ & 4 & Dynamic expert count \\
Top-$k$ selection & $k$ & 2 & Experts activated per token \\
Resolution levels & $R$ & 3 & Multi-resolution levels \\
Static expert depth & $L_{\text{static}}$ & 2 & Layers per static expert \\
Backbone depth & $L_{\text{dyn}}$ & 6& MoE-Mamba block count \\
Load balance weight & $\lambda$ & 0.01 & Balance loss coefficient \\
\bottomrule
\end{tabular}%
% }
\end{table}

\subsection{Complementary Roles of Static and Dynamic Experts}

The Figure~\ref{fig:scan}, originally presented in the Section~\ref{sec:scan} to motivate the design of selective scan, corresponds to the static and dynamic experts discussed here. We further extend this comparison by systematically contrasting the functional roles of static and dynamic experts in the Table.A~\ref{tab:role} below.

\begin{table}[]
\centering
\caption{Complementary Roles of Static and Dynamic Experts. }
\label{tab:role}
\begin{tabular}{lll} \hline
\textbf{Aspect} & \textbf{Static Experts}             & \textbf{Dynamic Experts}            \\ \hline
Routing         & Deterministic (resolution-based)    & Learned (content-based)             \\
Scope           & Resolution-specific                 & Shared across resolutions           \\
Stage           & Pre-scanning encoding               & Post-scanning modeling              \\
Purpose         & Resolution-aware feature extraction & Region-adaptive pattern recognition  \\
\bottomrule
\end{tabular}
\end{table}

Static experts address how tokens should be encoded at different resolutions, leveraging known structural priors from the multi-resolution preprocessing pipeline. By dedicating separate parameters to each resolution level, they can learn resolution-specific feature transformations without interference.

Dynamic experts address which contextual patterns should be emphasized within region-aware sequences, enabling adaptive modeling based on token content. By learning to route tokens to specialized experts, they can capture heterogeneous diagnostic patterns that may appear within or across regions, independent of resolution.

Together, static and dynamic experts decouple resolution-specific encoding from region-adaptive contextual modeling, enabling MoEMambaMIL to exploit both known structural priors and learned content-based specialization. This separation of concerns also facilitates interpretability: static expert assignments reveal resolution-level processing, while dynamic expert assignments reveal content-level specialization.

\subsection{Handling Non-Contiguous Token Sequences}
In practice, tokens routed to the same expert are not necessarily contiguous in the original scan order. Nevertheless, we adopt this setting based on both semantic intuition and empirical evidence. Since routing decisions are made according to token content, tokens assigned to the same expert tend to be semantically related despite being spatially dispersed. From this perspective, the Mamba layer serves as a contextual aggregator over semantically coherent token groups rather than strictly sequential inputs. Moreover, consistent with prior observations \citep{mambamil}, we find that this approximation leads to stable training dynamics and strong empirical performance, suggesting that the state-space model can effectively accommodate such non-contiguous inputs.

\subsection{Detailed Ablation Study}

This appendix provides a detailed analysis of the ablation studies Table~\ref{tab:analysis} summarized in the main paper.
We investigate the effects of (i) the number of MoE-Mamba layers, (ii) the number of activated experts per token, and (iii) the load balancing loss weight.
Unless otherwise specified, MoE modules are applied to every Mamba layer.

\paragraph{Network Depth.}
We vary the number of MoE-Mamba layers from 2 to 6 while fixing four experts per layer, activating two experts, and setting the load balancing coefficient $\lambda=0.001$.
As shown in Table~X, deeper architectures achieve higher macro-F1 and macro-MCC, indicating improved hierarchical modeling of complex pathological patterns.
Performance gains saturate at six layers, motivating our final depth choice.

\paragraph{Activated Experts.}
We analyze the impact of Top-$k$ expert activation under a fixed six-layer configuration.
Activating a single expert leads to inferior macro-F1 and macro-MCC, suggesting insufficient robustness due to excessive specialization.
In contrast, activating two experts yields the best overall performance.
Further increasing the number of activated experts gradually diminishes the benefits of sparse conditional computation, as the model approaches a dense mixture.

\paragraph{Load Balancing Loss Weight.}
We evaluate different values of the load balancing loss coefficient $\lambda$.
A small weight ($\lambda=0.001$) yields stable results, while larger values negatively affect macro-F1 and macro-MCC.
This suggests that overly strong load balancing constraints may restrict routing flexibility and limit effective expert specialization in heterogeneous pathological image data.

\subsection{Qualitative Analysis}
\label{appendix:Qualitative}

\begin{figure*}
    \centering
    \includegraphics[width=0.8\linewidth]{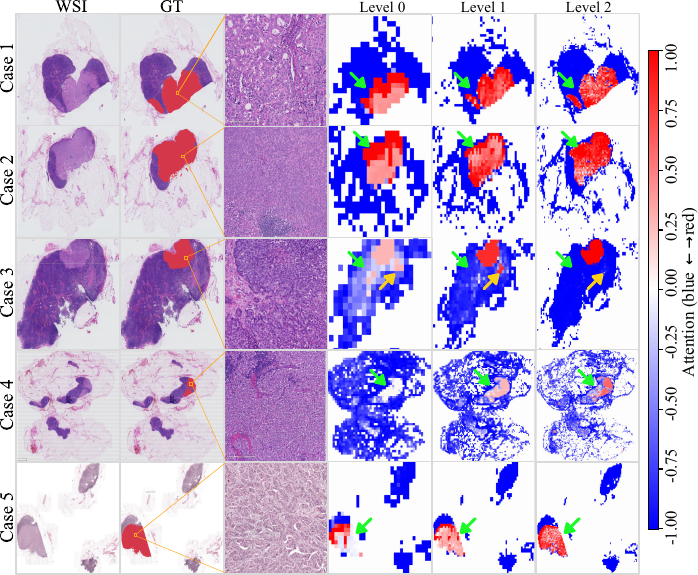}
    \caption{Multi-resolution MIL attention visualization.
    For each WSI, visualizing instance-level attention weights derived from the final MIL pooling for three resolution levels (Level 0–2). Red indicates high attention, blue indicates low attention, and white denotes intermediate importance.
}
    \label{fig:detailed_heatmap}
\end{figure*}

\subsubsection{Detailed Analysis of Multi-Resolution Attention}

Figure.A~\ref{fig:detailed_heatmap} provides a detailed visualization of instance-level attention weights produced by the final MIL pooling layer in MoEMambaMIL, separately for three resolution levels. Importantly, all attention maps are computed after the complete forward pass, including both static and dynamic expert processing, and therefore reflect each instance’s final contribution to the slide-level prediction.

\subsubsection{Consistent ROI Localization Across Resolutions}

Across all cases, instances from different resolutions consistently assign higher attention to regions overlapping with the ground-truth (GT) annotations. This demonstrates that the model does not rely on a single resolution for decision making, but instead learns resolution-invariant cues that align with diagnostically relevant regions.

\subsubsection{Resolution-Dependent Localization Precision}

Despite this consistency, the attention patterns exhibit clear resolution-dependent differences.  
In Case 1, Level 0 instances correspond to relatively large physical areas, limiting their ability to isolate small subregions within the GT. As a result, attention at Level 0 appears spatially diffuse. In contrast, Level 1 and Level 2 instances, with finer spatial granularity, successfully resolve and highlight the small GT subregion, yielding sharper and more localized attention maps.

\subsubsection{False Positive Suppression at High Resolution}

Case 3 illustrates the role of high-resolution instances in disambiguating visually similar regions. At coarse and intermediate resolutions (Level 0 and Level 1), the model assigns non-negligible attention to a secondary region outside the GT, reflecting structural similarity at larger scales. At Level 2, however, attention becomes tightly aligned with the true GT, while the spurious region is effectively suppressed. This suggests that fine-resolution instances capture discriminative morphological details that are not available at coarser scales.

\subsubsection{Progressive Evidence Refinement Across Scales}

In Case 4, GT regions receive intermediate attention at Level 0 (white regions), indicating uncertain but potentially relevant evidence. As resolution increases, attention on the same regions becomes progressively stronger (red at Level 1 and Level 2). This behavior reflects a refinement process in which coarse-scale instances provide weak evidence that is subsequently confirmed and strengthened by fine-scale observations.

\subsubsection{Summary}

Together, these observations indicate that different resolutions play complementary roles in the MIL framework: coarse resolutions favor broad localization and contextual awareness, while fine resolutions provide precise boundary delineation and false-positive rejection. By aggregating instance-level evidence across resolutions, MoEMambaMIL achieves robust and interpretable slide-level predictions.

%%%%%%%%%%%%%%%%%%%%%%%%%%%%%%%%%%%%%%%%%%%%%%%%%%%%%%%%%%%%%%%%%%%%%%%%%%%%%%%
%%%%%%%%%%%%%%%%%%%%%%%%%%%%%%%%%%%%%%%%%%%%%%%%%%%%%%%%%%%%%%%%%%%%%%%%%%%%%%%

\end{document}